\documentclass{article} % For LaTeX2e
\usepackage{iclr2021_conference,times}

\usepackage{amsmath,amsfonts,bm}

\def\eqref#1{equation~\ref{#1}}
\def\1{\bm{1}}

\DeclareMathAlphabet{\mathsfit}{\encodingdefault}{\sfdefault}{m}{sl}
\SetMathAlphabet{\mathsfit}{bold}{\encodingdefault}{\sfdefault}{bx}{n}

\DeclareMathOperator*{\argmin}{arg\,min}

\usepackage{hyperref}
\usepackage{url}
\usepackage{graphicx} 

\usepackage{amsmath}
\usepackage{wrapfig}
\usepackage{multirow}
\usepackage{gensymb} % for degree symbol in normal estimation evaluation metrics
\usepackage{physics} % for capital delta operator (bi-level optim)
\usepackage{colortbl} % color tables (ch 6)
\usepackage{xcolor} % color tables (ch 6)
\definecolor{lg}{gray}{0.9}
\usepackage{booktabs}
\usepackage{subcaption}

\usepackage[numbers]{natbib}

\title{Enabling Asymmetric Knowledge Transfer in Multi-Task Learning with Self-Auxiliaries}

\author{Olivier Graffeuille, Yun Sing Koh \& J\"{o}rg Wicker \\
Department of Computer Science\\
The University of Auckland \\
\texttt{ogra439@aucklanduni.ac.nz,\{y.koh,j.wicker\}@auckland.ac.nz} \\
\AND
Moritz Lehmann \\
Starboard Maritime Intelligence \\
\texttt{moritz.lehmann@starboard.nz} \\
}

\iclrfinalcopy % Uncomment for camera-ready version, but NOT for submission.
\begin{document}

\maketitle

\begin{abstract}
Knowledge transfer in multi-task learning is typically viewed as a dichotomy; positive transfer, which improves the performance of all tasks, or negative transfer, which hinders the performance of all tasks. In this paper, we investigate the understudied problem of asymmetric task relationships, where knowledge transfer aids the learning of certain tasks while hindering the learning of others. We propose an optimisation strategy that includes additional cloned tasks named self-auxiliaries into the learning process to flexibly transfer knowledge between tasks asymmetrically. Our method can exploit asymmetric task relationships, benefiting from the positive transfer component while avoiding the negative transfer component. We demonstrate that asymmetric knowledge transfer provides substantial improvements in performance compared to existing multi-task optimisation strategies on benchmark computer vision problems.
\end{abstract}

\section{Introduction}
Multi-Task Learning (MTL) models learn multiple tasks jointly to exploit shared knowledge between tasks and improve the performance of all tasks. Knowledge is transferred between tasks in deep MTL systems by sharing neural network parameters~\cite{MTL_LTB,MTL_factorisation} or feature representations~\cite{MTL_crossstitch,MTL_sluice} between tasks. Generally, it is assumed that if the tasks being learnt are related then the knowledge transfer will be beneficial for learning, while dissimilar tasks may result in negative transfer where the tasks' performance decreases.

This view of knowledge transfer in multi-task learning implicitly assumes that task relationships are symmetric. The performance of tasks trained together will either improve for all tasks if they are related or reduce for all tasks if they are not. However, recent empirical studies~\cite{MTL_task_groupings,MTL_task_groupings_efficient} of MTL systems reveal that MTL problems can accommodate \emph{asymmetric task relationships}, where training multiple tasks together can substantially improve the performance of one task while harming the performance of another. These results motivate techniques where knowledge transfer between tasks is directed rather than symmetrical, to disentangle positive and negative knowledge transfer between the tasks in opposite directions. We refer to this as \emph{asymmetric knowledge transfer}. Applied to problems with asymmetric task relationships, asymmetric knowledge transfer may enable MTL models to harness positive transfer while avoiding negative transfer.

However, modern multi-task learning models do not exploit asymmetric task relationships. MTL architectures are comprised of shared and task-specific parameters~\cite{MTL_PCGrad,MTL_auto_lambda,MTL_conflict_averse,MTL_graddrop,MTL_MGDA,MTL_MOO,MTL_pareto,MTL_IMTL,MTL_weight_uncertainty,MTL_adashare}, where parameters shared by multiple tasks share knowledge between all tasks which train them, and task-specific parameters avoid knowledge transfer altogether. While many methods have been developed for effective optimisation of deep learning systems by weighing or modifying learning gradients, these strategies are defined by symmetric operations which ignore task relationships ~\cite{MTL_PCGrad,MTL_auto_lambda,MTL_conflict_averse,MTL_graddrop,MTL_MGDA,MTL_MOO,MTL_pareto,MTL_IMTL,MTL_weight_uncertainty,MTL_adashare}. 
Whereas MTL optimisation strategies weigh task losses during training to balance trade-offs in performance between tasks~\cite{MTL_MGDA,MTL_MOO,MTL_attention,MTL_IMTL}, asymmetric knowledge transfer may instead circumvent this trade-off and improve performance of beneficial tasks without degrading the performance of others. 

We introduce Self-Auxiliary Asymmetric Learning (SAAL), the first multi-task optimisation method that enables asymmetric knowledge transfer in multi-task learning models, allowing them to exploit asymmetric task relationships during training. To asymmetrically transfer knowledge from a task $\mathcal{T}_1$ to another task $\mathcal{T}_2$ in an MTL model, the learning signal from $\mathcal{T}_1$ must apply to the modules of the MTL model which predict $\mathcal{T}_2$, but not vice versa. We achieve this by creating identical clones of tasks that insert additional learning signals during training but are discarded at inference. Inspired by Wang et al.~\cite{MTL_small_head}, we term these clone tasks \emph{self-auxiliaries}. During training, these self-auxiliaries use task-specific modules specific to other tasks to make predictions, such that their learning signal is applied to network parameters that define other task functions. By selecting which self-auxiliary tasks to include during training, we can induce any configuration of knowledge transfer, including asymmetric transfer.
Our approach can be considered a multi-task auxiliary learning framework. While auxiliary learning aims to optimise the performance of a single primary task by training it alongside a set of auxiliary tasks that are discarded after training, under our framework, all tasks learnt by the MTL model are primary tasks that can each use a number of self-auxiliaries to aid their learning.

\begin{figure}[t]
\centering
\includegraphics[width=0.9\textwidth]{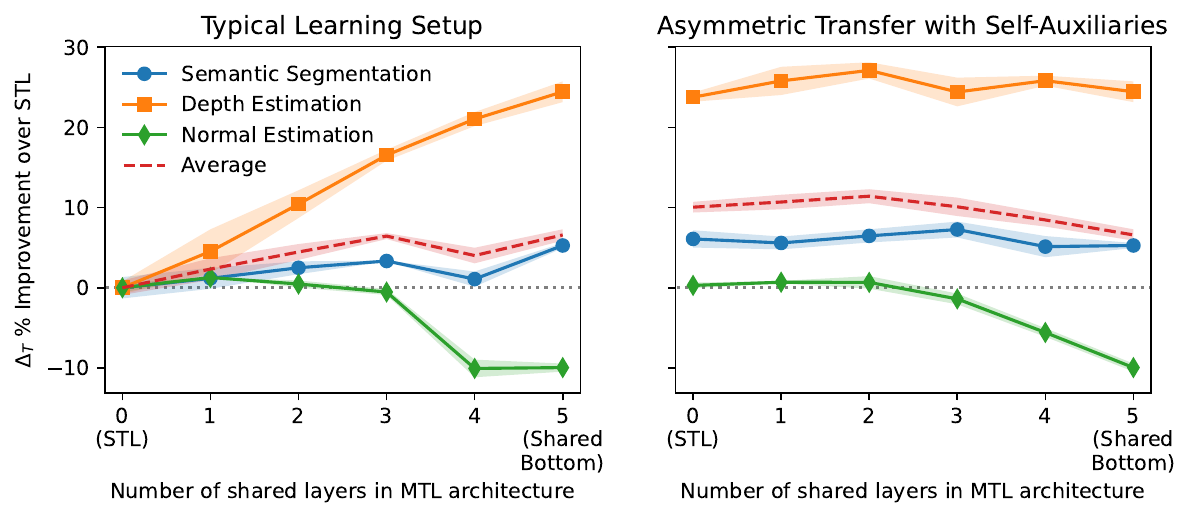}
%\caption{Performance of symmetric MTL vs. asymmetric MTL for NYUv2 dataset with varied number of shared layers.}
\caption[Performance of typical learning setup vs. asymmetric learning setup using pre-computed self-auxiliaries for the NYUv2 dataset with varied number of shared layers.]{Performance of typical learning setup vs. asymmetric learning setup using pre-computed self-auxiliaries for the NYUv2 dataset with varied number of shared layers, from no sharing (single-task learning models) to a fully shared encoder (shared bottom model).}
\label{diag asym example}
\end{figure}

We demonstrate the benefits of asymmetric transfer across different degrees of sharing in an MTL architecture for the NYUv2 dataset~\cite{dataset_nyuv2} in Figure~\ref{diag asym example}. 
We observe that under a typical learning setup that only supports symmetric knowledge transfer, as more layers are shared between tasks, the performance of depth estimation and semantic segmentation improve, but the performance of normal estimation degrades.
This indicates that knowledge transfer between these tasks hinders the learning of normal estimation but helps learn the other tasks, indicating asymmetric task relationships. Given this, we enable asymmetric transfer using self-auxiliaries to transfer knowledge from normal estimation to the other tasks but not vice versa. 
%We observe that this results in improved performance in semantic segmentation and depth estimation which is consistent across different numbers of shared layers, producing configurations with strong performance across all tasks without degradation of normal estimation when fewer than three shared layers.
The asymmetric MTL learning scheme maintains the benefits of knowledge transfer for depth estimation and semantic segmentation across sharing configurations, offering near-optimal performance on all tasks with 0 to 3 shared layers before normal estimation degrades in performance.
%We note that the improvement in performance from this asymmetric transfer are substantial, with an average improvement of 5\% with the optimal number of shared layers, which is [many times] greater than the gain achieved by state-of-the-art multi-task optimisation strategies on the same dataset~\cite{MTL_weight_random}. 
%However, this result relies on prior knowledge of asymmetric task relationships, which are typically not known.
% Brag about results here

The rest of this paper is structured as follows. In Section~\ref{asym related works}, we discuss related research on asymmetric transfer and task relationship learning in MTL. In Section~\ref{section asym method}, we formalise asymmetric task relationships and introduce our proposed SAAL framework for asymmetric knowledge transfer. In section~\ref{section asym experiments}, we evaluate the effectiveness of our approach. Section~\ref{section asym conclusion} concludes the paper.

\section{Related Work}\label{asym related works}
In this section, we review the most relevant existing research relating to asymmetric knowledge transfer and strategies for learning task relationships in multi-task learning.

\subsection{Asymmetric Transfer in Multi-Task Learning}
Asymmetric knowledge transfer is an understudied problem in MTL research~\cite{MTL_asymmetric,MTL_asymmetric2,MTL_asymmetric_application,MTL_asymmetric_gradient}.  
Gradsplit~\cite{MTL_asymmetric_gradient} is a multi-task optimisation approach that filters training signals from certain tasks to some parameter subgroups in a shared bottom MTL architectures based on task relationships which can be asymmetrical. This produces parameter subgroups within neural network layers learnt with asymmetric knowledge transfer. However, as each task model uses all task subgroups to make predictions, knowledge transfer still flows between all tasks through some network parameters. Furthermore, the final task model includes parameters that were not trained by that task. Finally, this approach is only compatible with shared bottom MTL architectures.
%Gradsplit~\cite{MTL_asymmetric_gradient} investigates asymmetric transfer in MTL optimisation. The shared model parameters are partitioned into $T$ groups each corresponding to a task, which act normally in the forward pass, but filter gradients in the backwards pass such that only gradients from tasks with a positive relationship to a task are applied to that task's group parameters. This produces subsets of each network module, which are learnt asymmetrically with regards to each task. However, this method is restricted to a Shared Bottom MTL architecture and is unable to learn task differences in the encoder, which is restrictive when aiming to optimise each task function with knowledge transfer. 
Auto-$\lambda$~\cite{MTL_auto_lambda} is another MTL optimisation algorithm that dynamically learns weights for task losses during training. Task relationships are inferred from relative task weights and are symmetrical in an MTL setting. However, these are used to indicate asymmetric task relationships when applied to multiple auxiliary learning models, which each use a single primary task. However, this approach can only support asymmetric task relationships in auxiliary learning settings, which avoid potential negative transfer as they only optimise the performance of the primary task.

Some research instead investigates MTL architectures to leverage asymmetric transfer, such as Asymmetric MTL (AMTL)~\cite{MTL_asymmetric}. AMTL uses a different network for each task and enforces that the parameters in each task network should approximately equal a linear combination of other task parameters using a regularisation term during training. Crucially, the coefficients that determine these linear combinations can be asymmetric, such that information flows from certain tasks to other tasks but not vice versa. These coefficients also directly indicate task relationships. However, directly transferring task parameters restricts the flexibility of each network to learn distinct features and transfers knowledge equally across all network layers. Furthermore, learning an entire task network per task scales poorly. 
Asymmetric Multi-Task Feature Learning (AFMTL)~\cite{MTL_asymmetric2} improves this approach by allowing network parameters to approximately equal a linear combination of bases rather than other task weights. For asymmetric knowledge transfer in this setting, an autoencoder loss term is added, which reconstructs the penultimate activation maps of each task from reliable task outputs. Transferring transfer in the feature space rather than between parameters allows this framework to be applied to various deep learning systems. 
Given an application with intuitive asymmetric task relationships, one work develops AMTA-Net~\cite{MTL_asymmetric_application} for semantic segmentation of prostate beds in CT scans, where knowledge is transferred asymmetrically from highly relevant but simpler tasks. This work relies on predefined task relationships and an auxiliary learning framework where a single primary task is optimised and so does not consider potential negative transfer towards auxiliary tasks caused by asymmetric relationships.

\subsection{Task Relationship Learning}
How to determine task relationships is an important open problem in multi-task learning. Learning task relationships can improve modelling by informing which tasks should be learnt together~\cite{MTL_task_groupings}, how to transfer knowledge between tasks~\cite{MTL_MRN,MTL_sluice} and offer an understanding of the problems we are modelling. Some approaches represent task relationships in a matrix of size $T \times T$ where each element indicates the quantity of knowledge transfer between the tasks, and optimise this matrix during training~\cite{MTL_MRN,MTL_crossstitch,MTL_sluice,MTL_principled}. Many other MTL architectures jointly train multiple tasks and automatically determine what knowledge to share between tasks without explicitly learning task relationships~\cite{MTL_attention,MTL_LTB,MTL_NAS,MTL_adashare}. 

Some works including Gradsplit~\cite{MTL_asymmetric_gradient} avoid the challenge of learning task relationships by performing complete enumeration, optimising a model with each combination of tasks to determine positive and negative relationships. An empirical study of multi-task learning on vision tasks also performed complete enumeration to analyse the impact of tasks on different tasks' performance, revealing several key findings~\cite{MTL_task_groupings}. Most importantly, they find that task relationships are asymmetric, in that training multiple tasks together can improve the performance of one task while reducing the performance of another. 
Task relationships depend on many factors, including the model architecture and dataset size. Finally, MTL relationships are only weakly related to transfer relationships whereby a model is trained on a task and fine-tuned on another~\cite{MTL_taskonomy}. However, this complete enumeration approach to learning task relationships for MTL scales quadratically with the number of tasks being modelled and is often not computationally viable. One work only computes the performance of a subset of possible task combinations and learns a model to generalise to unknown task relationships~\cite{MTL_negative_transfer}. 

Other works develop heuristics to estimate task relationships with reduced computational cost. Common strategies rely on look-ahead losses~\cite{MTL_auto_lambda,MTL_task_groupings_efficient,MTL_gradient_coordination}, gradient conflicts between tasks~\cite{MTL_olaux,MTL_recon,MTL_conal,MTL_rotograd,MTL_conflict_averse} or feature similarity~\cite{MTL_taskonomy,MTL_task_relationship_RSA,MTL_task_relationship_DDS,MTL_task_relationship_attribution}, although other domain-specific strategies exist for reinforcement learning~\cite{MTL_robot_relationships} and evolutionary multitasking~\cite{MTL_task_relatedness_evolution}.
Look-ahead loss strategies take a virtual training step with a task and analyse how this affects the performance of other tasks. One work uses look-ahead losses to approximate task relationships and propose task groupings more efficiently than complete enumeration~\cite{MTL_task_groupings_efficient}. A similar strategy is to compute the angle between the learning gradients of each task, where an angle of less than $\frac{\pi}{2}$ is typically interpreted as positive transfer between highly related tasks, while the tasks are considered to be conflicting otherwise~\cite{MTL_PCGrad,MTL_conflict_averse}. This approach is also used to determine MTL architectures that minimise negative transfer~\cite{MTL_recon,MTL_conal,MTL_rotograd}. Some works show that gradient angle is approximately equivalent to look-ahead loss~\cite{MTL_recon,MTL_gradient_coordination}. Both strategies provide an approximation of task relationships in the local search space of a neural network while it trains. The relationships found by these approaches are also symmetrical by definition, as the angle between vectors is a commutative function.
Feature similarity metrics aim to compare learnt network weights~\cite{MTL_task_relationship_attribution} or activation maps~\cite{MTL_task_relationship_RSA,MTL_task_relationship_DDS} between task data or models to estimate relationships. One uses a single-task network trained on each task to generate deep representations for a set of images and uses Representation Similarity Analysis (RSA) to establish task relationships~\cite{MTL_task_relationship_RSA}. Another approach investigates a range of similarity functions over activation maps of a batch of images~\cite{MTL_task_relationship_DDS}. Instead of comparing learnt representations, another approach investigates the attribution maps of the neural network weights of single-task networks given the same input data~\cite{MTL_task_relationship_attribution}. 
% What is constant across these methods is that the only approach to determining asymmetric task relationships is to (either asym MTL methods lol) or to train networks to optimality, local task relatinoships method are symmetric.

\section{Self-Auxiliary Asymmetric Learning}\label{section asym method}
Consider a multi-task learning model $g$ partitioned into neural network parameters shared by all tasks $\theta_{sh}$ and task-specific parameters $\{\theta_1, \theta_2, \ldots, \theta_T\}$ such that task $\mathcal{T}_t$ is modelled by $g_t(\textbf{x} \, | \, \theta_{sh}, \theta_t)$. 

\paragraph{Asymmetric task relationships.} Let the final performance of $g$ on $\mathcal{T}_t$ after being trained trained on tasks $\{\mathcal{T}_1, \mathcal{T}_2, \ldots\}$ be given by $A_t^{\{1,2,\ldots\}}$.
Symmetric task relationships imply that all task relationships take the form of either $(A_1^{\{1,2\}} > A_1^{\{1\}}) \land (A_2^{\{1,2\}}> A_2^{\{2\}})$ in the case of positive transfer or $(A_1^{\{1,2\}} < A_1^{\{1\}}) \land (A_2^{\{1,2\}} < A_2^{\{2\}})$ in the case of negative transfer.
However, asymmetric task relationship can exist~\cite{MTL_task_groupings}, where for example $(A_1^{\{1,2\}} > A_1^{\{1\}}) \land (A_2^{\{1,2\}} < A_2^{\{2\}})$ if $\mathcal{T}_1$ improves the performance of $\mathcal{T}_2$ but $\mathcal{T}_2$ reduces the performance of $\mathcal{T}_1$~\cite{MTL_task_groupings}. %or vice versa 
We aim to exploit asymmetric task relationships by enabling asymmetric knowledge transfer in MTL, where tasks benefit from positive transfer but prevent negative transfer between the same tasks.

%Given the desirable property that the training signal from a task affects all parameters which define its task function, components of the MTL model, which are shared between multiple tasks, will be trained by each task. 
Modern MTL architectures combine shared and task-specific neural network parameters. The knowledge transfer that occurs in shared components is symmetric as each task trains neural network parameters that are used by all task functions, such that the training from each task affects the performance of all other tasks. Task-specific network components, however, only affect the performance of that task, and so may support asymmetric knowledge transfer if knowledge from other tasks is selectively transferred to these components without affecting other task functions.
% however, current approaches do not
%To do so, an arbitrary task $\mathcal{T}_t$, should learn from all tasks which benefit it, but not tasks which hinder its learning. 
%An arbitrary task $\mathcal{T}_t$ which is modelled by $g_t$ which contains both shared and task-specific components may benefit from knowledge transfer from certain tasks. 
%This requires this task function $g_t$ to be trained by other tasks as well as $\mathcal{T}_t$. These other tasks use task functions which typically only share some neural network components in common with $g_t$. 

\begin{figure}[t]
\centering
\includegraphics[width=0.7\textwidth]{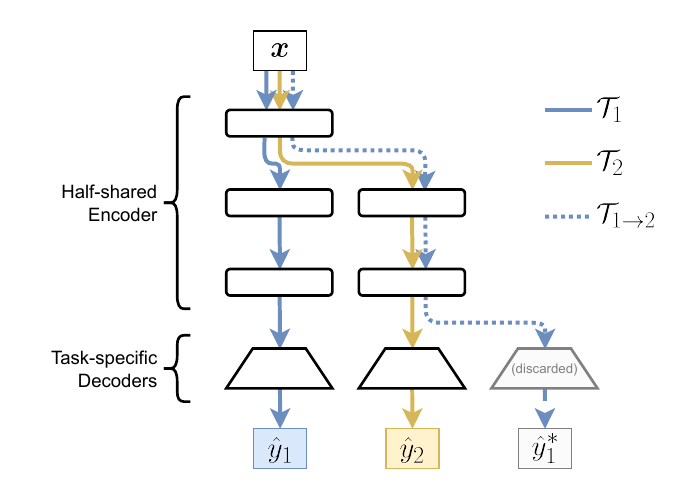}
\caption[Conceptual diagram for a self-auxiliary task, in a branching multi-task architecture with two tasks.]{Conceptual diagram for a self-auxiliary task, in a branching multi-task architecture with two tasks. These tasks have an asymmetric relationship where $\mathcal{T}_1$ improves the learning of $\mathcal{T}_2$ but $\mathcal{T}_2$ impairs the learning of $\mathcal{T}_1$.
To enable asymmetric transfer, a self-auxiliary $\mathcal{T}_{1 \to 2}$ is added. It is identical to its source task $\mathcal{T}_1$ and uses the modules of its target task $\mathcal{T}_2$ to share knowledge ($\mathcal{T}_{1 \to 2}
\leftrightarrow \mathcal{T}_2$), and is discarded at inference. This induces positive transfer in task-specific layers of the $\mathcal{T}_2$ model ($\mathcal{T}_1 \to \mathcal{T}_2$) while avoiding negative transfer in task-specific layers of the $\mathcal{T}_1$ model ($\mathcal{T}_2 \not\to \mathcal{T}_1$).}
%To enable asymmetric transfer, a self-auxiliary $\mathcal{T}_{1 \to 3}$ shares modules with task $\mathcal{T}_3$ to share knowledge $\mathcal{T}_1 \to \mathcal{T}_3$, but $\mathcal{T}_1$ learns independently of $\mathcal{T}_3$ after the first layer such that knowledge does not flow $\mathcal{T}_3 \not\to \mathcal{T}_1$.
\label{fig:diag self auxiliary}
\end{figure}

\paragraph{Method overview.}
To leverage the training signal from other tasks in task-specific components, we create task clones termed self-auxiliaries and use these to train the task-specific components of other tasks. A self-auxiliary task that clones another task $\mathcal{T}_1$ can train $\theta_2$, transferring knowledge from $\mathcal{T}_1$ to $\mathcal{T}_2$ without affecting the task-specific parameters $\theta_1$, as illustrated in Figure~\ref{fig:diag self auxiliary}. These self-auxiliaries therefore offer a method for directed knowledge transfer between tasks, and many self-auxiliaries can be created to implement any configuration of knowledge transfer, including asymmetric transfer. Self-auxiliaries are discarded when training is completed.
%To leverage the training signal from other tasks in task-specific components $\theta_t$, we create clones of other tasks termed self-auxiliaries, and use these to train $g_t$. A self-auxiliary task that clones another task $\mathcal{T}_s$ can train $\theta_t$, transferring knowledge from $\mathcal{T}_s$ to $\mathcal{T}_t$ without affecting the task-specific parameters $\theta_s$. These self-auxiliaries therefore offer a method for directed knowledge transfer between tasks, and many self-auxiliaries can be created to implement any configuration of knowledge transfer, including asymmetric transfer. Self-auxiliaries are discarded when training is completed.

Training neural networks with self-auxiliaries, which are discarded at inference, also separates two effects of task-specific components in MTL architectures: added representation capacity and interacting training signals. 
%Task-specific components in MTL architectures provide two effects. 
Task-specific components increase the representation capacity of models to flexibly learn a task's structure, enabling richer learning of task differences. They also split task learning signals, preventing both positive and conflicting interactions from the other tasks as they do not use the module. 
By using self-auxiliaries to train task-specific components of other tasks, we benefit from additional representation capacity while still learning from other tasks by enabling knowledge transfer in task-specific modules. If the learning signal enabled by self-auxiliaries produces positive transfer, this will improve learning in these task-specific modules.

In the following sections, we describe our implementation of self-auxiliaries, then describe strategies for determining task relationships to select self-auxiliary tasks that promote positive asymmetric knowledge transfer during training.

%The factors that determine transfer relationships are not well understood and multi-faceted~\cite{}. Potential explanations for asymmetric transfer may include one task having noisier data than another, one task learning more informative labels than another. [Some example about if there exists a process A -> B -> C, the model A -> B may help A -> C but not vice versa?]

\subsection{Self-auxiliaries}
%somewhere: say simple

Alongside the existing tasks we aim to optimise $\{\mathcal{T}_1, \mathcal{T}_2, \ldots, \mathcal{T}_T\}$ we introduce self-auxiliary tasks $\{\mathcal{T}_{s \to t} \, | \, s, t \in \{1, 2, \ldots, T\}, s \neq t\}$ which each transfer knowledge from a \emph{source} task to a \emph{target} task. 

We assume that the MTL model $g$ consists of an encoder-decoder architecture. The encoder consists of both shared and task-specific components, while the decoders are task-specific, as is typical of recent MTL architectures~\cite{MTL_adashare,MTL_attention,MTL_maximum_roaming,MTL_LTB,MTL_MMoE,MTL_NAS}. Self-auxiliary tasks use identical data, labels, data augmentations, and other training implementations as their source task, but use the same encoder as their target task to transfer knowledge to the target task's task-specific components, as illustrated in Figure~\ref{fig:diag self auxiliary}. Self-auxiliaries use independent decoders to flexibly learn shared representations within the shared encoder. 

As self-auxiliaries are not computed at inference, these additional encoders are discarded after training. These self-auxiliaries, therefore, do not add any model parameters or computations to the final MTL model.
In cases where multiple primary tasks use identical encoders, as may occur in neural architecture search-based architectures~\cite{MTL_NAS,MTL_LTB,MTL_adashare}, these tasks share self-auxiliaries, reducing the number of self-auxiliaries. For simplicity, we omit these cases from our formulation. 

We note that this implementation results in duplicate learning signals from source tasks and self-auxiliaries which clone them to affect shared modules, such as the first layer of the network in Figure~\ref{fig:diag self auxiliary} which is trained by both $\mathcal{T}_{1}$ and $\mathcal{T}_{1 \to 2}$. Empirically, we find that nullifying duplicate learning signals from self-auxiliaries or including new modules for self-auxiliaries to train separately decreased model performance.

\subsection{Determining task relationships}
Our implementation of self-auxiliaries can transfer knowledge from any task to any other task in the model during training. This enables asymmetric transfer between task pairs. In this section, we introduce strategies to determine asymmetric task relationships, to inform which self-auxiliaries to include to include in the learning process. The aim of these strategies is to select directed task transfers to include in the learning process to maximise positive transfer and avoid negative transfer. 

We introduce coefficients $\omega$, which represent the inclusion of primary tasks and self-auxiliaries into model training. Let each coefficient $\omega_t$ be associated with a primary task $\mathcal{T}_t$, and $\omega_{s \to t}$ be associated with a self-auxiliary task $\mathcal{T}_{s \to t}$. These coefficients represent the importance of each task for training. Tasks are included in training if their corresponding coefficient $\omega > 0$. The optimal task coefficients are given by the following bi-level optimisation problem~\cite{NAS_DARTS}:
\begin{align} 
    \min_{\omega} \hspace{0.5em} &\mathcal{L}^{\text{val}}(\theta^*(\omega))\label{equation bi-level 1} \\ 
    \text{s.t.} \hspace{0.5em} &\theta^*(\omega) = \argmin_{\theta} \mathcal{L}^{\text{train}}(\theta,\omega)\label{equation bi-level 2}
\end{align}
where
\begin{align}
    \mathcal{L}^{\text{val}}(\theta^*(\omega)) &= \sum_{t=0}^T L_t(f(\boldsymbol{x}^{\text{val}} \, | \, \theta_{sh}^*(\omega), \theta_{t}^*(\omega)), y_t^{\text{val}})\\
    \mathcal{L}^{\text{train}}(\theta,\omega) &= \sum_{t=0}^T \left( \omega_t L_t(f(\boldsymbol{x}_t^{\text{train}} \, | \, \theta_{sh}, \theta_{t}), y_t^{\text{train}})  + \sum_{\substack{s=0 \ s \neq t}}^T \omega_{s \to t} L_s(f(\boldsymbol{x}_s^{\text{train}} \, | \, \theta_{sh}, \theta_{t}), y_s^{\text{train}}) \right)
\end{align}
where $L_t$ represents the loss function for task $\mathcal{T}_t$ and the training and validation data for task $\mathcal{T}_t$ are represented by $(\boldsymbol{x}_t^{\text{train}}, y_t^{\text{train}})$ and $(\boldsymbol{x}_t^{\text{val}}, y_t^{\text{val}})$, respectively. This optimisation problem aims to find coefficients $\omega$, which when applied to a MTL model and trained to optimality $\theta^*$ using self-auxiliaries, produce the best performance with primary tasks on the validation set. The upper-level problem optimises for validation loss, to avoid overfitting $\omega$ to the training data. The validation loss excludes self-auxiliaries as we do not aim to maximise their performance at inference, and weighs the loss of all primary tasks equally without task coefficients to avoid degenerate solution $\omega = 0$.  

This optimisation problem is computationally intractable. We propose two strategies and a combined strategy to select task coefficients $\omega$ for this learning problem.

\paragraph{Enumeration strategy.}
We evaluate the performance of a fully trained Shared Bottom MTL model with every task pair to establish definitive task relationships, similarly to other works~\cite{MTL_task_groupings,MTL_asymmetric_gradient}. Task coefficients $\omega$ produced by this strategy are booleans, which determine whether each task is included in training:
\begin{equation}
    \omega_t = 1, \qquad
    \omega_{s \to t} = \begin{cases}
        1 & \text{if} \hspace{0.5em} A_t^{\{s,t\}} > A_t^{\{t\}} \\
        0 & \text{otherwise}. 
    \end{cases}
\end{equation}
This strategy selects self-auxiliaries beneficial to the learning of their target task to be included during the training of the MTL model, along with the primary tasks. The task relationships produced by this approach are asymmetric, and implementing these relationships with self-auxiliaries will therefore introduce asymmetric knowledge transfer. However, this approach is computationally expensive and scales quadratically with the number of tasks in the dataset.

\paragraph{Loss weighting strategy.}
This approach learns the relative importance of primary and auxiliary tasks during model learning. By learning large coefficients for tasks that provide positive knowledge transfer and small coefficients for tasks that induce negative transfer, this approach can maximise positive transfer of self-auxiliaries towards their target task. To do so, we solve an approximation of the bi-level optimisation problem in Equations~\ref{equation bi-level 1}-~\ref{equation bi-level 2}, following established methodologies~\cite{NAS_DARTS,MTL_auto_lambda}. We simplify the upper-level optimisation problem, firstly by approximating the optimal network parameters $\theta^*(\omega)$ with a virtual training step and then applying the chain rule:
\begin{align}
    &\grad_{\omega} \mathcal{L}^{\text{val}}(\theta^*(\omega)) \\
    \approx \, &\grad_{\omega} \mathcal{L}^{\text{val}}(\theta^{\prime}), \qquad \theta^{\prime} = \theta - \eta \grad_{\theta} \mathcal{L}^{\text{train}}(\theta,\omega) \\
    = \, &\grad_{\omega}\mathcal{L}^{\text{val}}(\theta^{\prime}) - \eta \grad_{\omega, \theta}^2 \mathcal{L}^{\text{train}}(\theta, \omega) \grad_{\theta^{\prime}}\mathcal{L}^{val}(\theta^{\prime}).\label{equation chain rule}
\end{align}
The resulting term has second-order gradient terms which are difficult to optimise in neural networks. Given a small constant scalar $\epsilon$, Equation~\ref{equation chain rule} can be simplified by finite different approximation:
\begin{align}
    \approx \, &\frac{\grad_{\omega} \mathcal{L}^{\text{train}}(\theta^+,\omega) - \grad_{\omega} \mathcal{L}^{\text{train}}(\theta^-,\omega)}{2\epsilon} 
\end{align}
where $\theta^{\pm} = \theta \pm \epsilon \grad_{\theta^{\prime}}\mathcal{L}(\theta^{\prime},\omega)$. We set $\epsilon$ to the model learning rate $\eta$ for all experiments~\cite{MTL_auto_lambda}. This term includes first-order gradients, which can be optimised by any neural network gradient optimiser to learn values for $\omega$, requiring two forward passes and two backward passes. Task coefficients $\omega$ are uniformly initialised at the start of training.

In traditional MTL setups, approaches to learning coefficients for task losses act to balance the relative importance of task learning signals for model training~\cite{MTL_gradnorm,MTL_attention,MTL_auto_lambda,MTL_weight_random}. Instead, the coefficient of a self-auxiliary task represents the importance of knowledge transfer from its source task to its target task. As such, these coefficients represent directed task relationships between primary tasks. Similarly to the enumeration strategy, applying these coefficients to self-auxiliaries can enable asymmetric knowledge transfer.

An advantage of this strategy is that task coefficients can be learnt in a single training run of the model. The coefficients are also continuous rather than discrete, offering more fine-grained task relationships, and are dynamic, allowing them to change at different stages of model training. However, they provide less reliable transfer relationships as they rely on local heuristics to learn task coefficients, and do not eliminate negative transfer completely as harmful tasks maintain non-zero coefficients.
% # forward / backward passes 

\paragraph{Combined strategy.}
We additionally combine the enumeration and loss weighting strategies to exploit the benefits of both approaches. Let the task coefficients produced by the enumeration strategy be $\omega^e$, and the normalised coefficients produced by the loss weighting strategy be $\omega^w$. We combine these coefficients by computing their product to produce combined coefficients $\omega^{ew}$:
\begin{equation}
    \omega_t^{ew} = \omega_t^{e} \omega_t^{w}, \qquad
    \omega_{s \to t}^{ew} = \omega_{s \to t}^{e} \omega_{s \to t}^{w}.
\end{equation}
This strategy is equivalent to the loss weighting strategy but additionally excludes self-auxiliaries that produce negative transfer. 
We refer to our Self-Auxiliary Asymmetric Learning algorithm with each of these task relationship strategies as \textbf{SAAL$_\text{e}$}, \textbf{SAAL$_\text{w}$} and \textbf{SAAL$_\text{ew}$}, respectively.

\paragraph{Normalisation.}
We normalise the task coefficients to balance the effect of each primary task on the overall training process, giving us the final task coefficients:
\begin{align}
    \bar{\omega_t} = \frac{\omega_t}{\omega_t + \sum_{\substack{s=0 \ s \neq t}}^T \omega_{s \to t}}, \qquad \bar{\omega}_{s \to t} = \frac{\omega_{s \to t}}{\omega_t + \sum_{\substack{s=0 \ s \neq t}}^T \omega_{s \to j}}.
\end{align}

\section{Experiments}\label{section asym experiments}
% robustness to noise task exp (like Auto-$\lambda$ does i think?)

We evaluate the effectiveness of Self-Auxiliary Asymmetric Learning for multi-task optimisation on established computer vision MTL benchmarks.

% why are we using CV?

\paragraph{Baseline methods.}
%something about architectures
We compare the performance of our approach to other MTL optimisation methods. We include task weighting MTL optimisation methods \textbf{Equal} weighting, \textbf{Uncertainty} Weighting~\cite{MTL_weight_uncertainty}, \textbf{DWA} (Dynamic Weight Averaging)~\cite{MTL_attention} and \textbf{Auto-$\lambda$}~\cite{MTL_auto_lambda}, as well as gradient manipulation methods~\textbf{PCGrad} (Projecting Conflicting Gradients)~\cite{MTL_PCGrad}, \textbf{CAGrad} (Conflict Averse Gradients)~\cite{MTL_conflict_averse} and \textbf{GradDrop}~\cite{MTL_graddrop}.

\paragraph{Relative task improvements.}
When displaying results, we include the relative task improvement compared to a single task learning baseline for each task~\cite{MTL_rotograd,MTL_recon,MTL_conal}. The individual relative improvement of a task and the average relative improvement over all tasks, respectively are given as:
\begin{equation}
\Delta_{\mathcal{T}_t} = 100\% \times \frac{1}{m_t} \sum_{i=1}^{m_t} (-1)^{l_i} \frac{M_i - S_i}{S_i}, \qquad \Delta_{MTL} = \frac{1}{T} \sum_{t=1}^{T} \Delta_{\mathcal{T}_t}
\end{equation}
where $m_t$ indicates the number of metrics for task $\mathcal{T}_t$, performance scores for the MTL model and STL baseline according to the $i^{\text{th}}$ metric for that task are given by $M_i$ and $S_i$ respectively and $l_i=1$ if a lower value is better for that metric and $0$ otherwise. For clarity, in our experiments we indicate tasks by their first letter, \emph{e.g.} $\Delta_S$ indicates the relative task performance for Semantic segmentation~\cite{MTL_rotograd}.

\paragraph{Network architectures.}
We evaluate each optimisation method on a half-shared MTL architecture where shallow layers are shared between all tasks and deeper layers are task-specific. Regarding the computer vision datasets, we share the first three blocks of the ResNet CNN architecture~\cite{resnet} and allow the last two to be task-specific. We also include STL and Shared Bottom architectures as baselines, which share no parameters and all parameters in the encoder between tasks, respectively.

\paragraph{Evaluation setup.}
Computer vision experiments were repeated over three seeds, with average results displayed. All methods used an initial learning rate $\eta = 10^{-1}$ and a cosine learning rate scheduler~\cite{MTL_auto_lambda}, where the number of epochs was tuned for each dataset. The model checkpoint with the best validation relative task improvement was selected for evaluation. Regarding learning of $\omega$ for SAAL$_\text{w}$ and SAAL$_\text{ew}$, we use an Adam~\cite{optim_adam} optimiser with learning rate $10^{-4}$. All methods used default hyperparameters with no tuning. 

%\subsubsection{Experimental Setup}

\paragraph{Datasets.}
We evaluated the effectiveness of our approach on two datasets with dense prediction tasks describing visual scenes, and a multi-task classification dataset. 
The \textbf{NYUv2 dataset}~\cite{dataset_nyuv2} has three tasks: 13-class semantic segmentation, depth prediction, and surface normal prediction. 
The \textbf{Cityscapes dataset}~\cite{dataset_cityscapes} has three tasks: 19-class semantic segmentation, 10-class part segmentation~\cite{dataset_cityscapes2} and disparity (inverse depth) estimation. We follow existing experimental setups~\cite{MTL_weight_uncertainty,MTL_attention,MTL_auto_lambda} with a DeepLabV3~\cite{deeplabv3} architecture with a ResNet-50~\cite{resnet} backbone for these two datasets. 
The \textbf{CelebA dataset}~\cite{dataset_celeba} describes boolean attributes about the faces of celebrities. We use a subset of nine tasks following existing experimental setups~\cite{MTL_task_groupings_efficient,MTL_conal} with a ResNet-18~\cite{resnet} backbone. Due to computation constraints and to allow for multiple runs to improve the reliability of results, we use 10\% of the 200k training images. Both dense prediction datasets were trained over 200 epochs~\cite{MTL_attention} while CelebA was trained over 25.

\subsection{Results}
Average results are displayed in Tables~\ref{table nyuv2} to \ref{table celeba}. On the NYUv2 dataset, all methods improved performance for depth estimation, most methods improved performance for semantic segmentation, but all methods reduced performance for normal estimation, indicating asymmetric task relationships between these tasks. 
%The only method to improve performance on normal estimation was our SAAL$_\text{ew}$ method, indicating that it avoided negative transfer towards normal estimation. 
Overall, SAAL$_\text{ew}$ substantially outperformed all existing MTL optimisation algorithms, and SAAL$_\text{e}$ achieved the second-best performance. These methods enhance positive transfer towards semantic segmentation and depth estimation while mitigating negative transfer towards normal estimation. SAAL$_\text{w}$ outperforms all existing methods for depth estimation but performs poorly on normal estimation. This indicates that the task weighting strategy is insufficient to prevent the negative transfer towards this task. The learning strategy that best minimised negative transfer towards normal estimation was Equal weighting, indicating that other approaches may aggravate this negative transfer.

Results on the Cityscapes dataset also indicate asymmetric relationships, with all approaches outperforming independent STL models on disparity prediction, but performing worse for semantic segmentation and part segmentation. Overall, SAAL$_\text{e}$, SAAL$_\text{ew}$ and SAAL$_\text{w}$ achieved the best, second-best and fourth-best performance, respectively. SAAL$_\text{e}$ was the most effective optimisation method in mitigating negative transfer towards semantic segmentation but was outperformed by a Shared Bottom architecture, while Uncertainty weighting best mitigated negative transfer towards part segmentation. 

The CelebA dataset also demonstrates asymmetric relationships, as most methods achieve higher accuracy than STL baselines for A3, A7 and A9 and lower accuracy for A2, A4 and A8. The inconsistent transfer towards the other tasks, A1, A5 and A6 may be influenced by high baseline accuracy for these tasks. Overall, SAAL$_\text{w}$, SAAL$_\text{ew}$ and SAAL$_\text{e}$ achieved the best, second-best and fourth-best performance, respectively.
%This indicates that this strategy is viable for MTL problems with approximately this many tasks.

\begin{table}
\centering
\caption[Performance comparison of SAAL and baseline methods for the NYUv2 dataset.]{Performance comparison of SAAL and baseline methods for the NYUv2 dataset. The MTL method with the best relative task improvement is bolded. Individual task metrics that outperform the STL baseline are highlighted in grey.}\label{table nyuv2}
\setlength{\tabcolsep}{4pt}
\fontsize{6pt}{7pt} \selectfont
\begin{tabular}{ll|rrrr|rrrrrrrrr}
\toprule
\multicolumn{1}{l}{}                   &                           & \multicolumn{4}{c}{Relative Improvement $\uparrow$} & \multicolumn{2}{c}{\textbf{S}emantic Seg. $\uparrow$} & \multicolumn{2}{c}{\textbf{D}epth $\downarrow$} & \multicolumn{5}{c}{\textbf{N}ormal} \\
\cmidrule(lr){3-6}
\cmidrule(lr){7-8}
\cmidrule(lr){9-10}
\cmidrule(lr){11-15}
& & \multicolumn{4}{c}{ }  & & & \multicolumn{2}{c}{ } & \multicolumn{2}{c}{Angle Distance $\downarrow$} &  \multicolumn{3}{c}{Within $t\degree\uparrow$} \\
%\cmidrule(l){9-10}
\cmidrule(lr){11-12}
\cmidrule(lr){13-15}
\multicolumn{1}{l}{Architecture}       & Method        & $\Delta_{\text{S}}$ & $\Delta_{\text{D}}$ & $\Delta_{\text{N}}$ & $\Delta_{\text{MTL}}$ & mIoU     & Pix acc    & Abs err & Rel err & Mean err & Med err & 11.25 & 22.5  & 30                   \\
\midrule
STL                            & -                                         & 0.00             & 0.00             & 0.00              & 0.00             &               38.05 &               57.44 &               60.64 &               24.81 &  22.32 & 15.41 & 41.00 & 66.15 & 74.95 \\
\midrule
\multirow{10}{*}{Half-shared}  & Equal                                     & 3.33             & 16.53            & \textbf{-0.54}    & 6.44             & \cellcolor{lg}39.21 & \cellcolor{lg}59.52 & \cellcolor{lg}50.67 & \cellcolor{lg}20.68 &  22.48 & 15.43 & 40.85 & 65.64 & 74.46 \\
                               & Uncert.~\cite{MTL_weight_uncertainty} & 4.47             & 19.74            & -5.06             & 6.39             & \cellcolor{lg}40.97 & \cellcolor{lg}58.18 & \cellcolor{lg}49.39 & \cellcolor{lg}19.62 &  23.70 & 16.02 & 38.23 & 62.90 & 72.37 \\
                               & DWA~\cite{MTL_attention}                  & 2.59             & 17.69            & -5.51             & 4.92             & \cellcolor{lg}39.98 & \cellcolor{lg}57.51 & \cellcolor{lg}50.81 & \cellcolor{lg}20.05 &  23.81 & 16.11 & 37.95 & 62.70 & 72.21 \\
                               & Auto-$\lambda$~\cite{MTL_auto_lambda}     & -4.67            & 18.53            & -1.84             & 4.01             & 34.63               & 57.24               & \cellcolor{lg}49.55 & \cellcolor{lg}20.15 &  22.72 & 15.77 & 40.14 & 65.02 & 74.02 \\
                               & PCGrad~\cite{MTL_PCGrad}                  & -0.12            & 16.10            & -6.63             & 3.12             & \cellcolor{lg}38.46 & 56.68               & \cellcolor{lg}51.42 & \cellcolor{lg}20.59 &  24.09 & 16.37 & 37.49 & 62.14 & 71.69 \\
                               & CAGrad~\cite{MTL_PCGrad}                  & 3.69             & 18.17            & -4.61             & 5.75             & \cellcolor{lg}40.53 & \cellcolor{lg}57.93 & \cellcolor{lg}50.13 & \cellcolor{lg}20.09 &  23.64 & 15.91 & 38.44 & 63.16 & 72.56 \\
                               & GradDrop~\cite{MTL_conflict_averse}       & 1.32             & 14.45            & -6.68             & 3.03             & \cellcolor{lg}39.32 & 57.04               & \cellcolor{lg}52.24 & \cellcolor{lg}21.08 &  24.07 & 16.39 & 37.40 & 62.10 & 71.72 \\
                               & SAAL$_\text{e}$                           & 7.26             & 24.41            & -1.39             & 10.09            & \cellcolor{lg}41.29 & \cellcolor{lg}60.89 & \cellcolor{lg}46.02 & \cellcolor{lg}18.68 &  23.37 & 15.66 & 38.99 & 63.71 & 73.09 \\
                               & SAAL$_\text{w}$                           & 3.06             & 23.79            & -13.26            & 4.53             & \cellcolor{lg}40.15 & \cellcolor{lg}57.78 & \cellcolor{lg}47.55 & \cellcolor{lg}18.36 &  25.27 & 18.08 & 33.62 & 58.88 & 69.41 \\
                               & SAAL$_\text{ew}$                          & \textbf{8.68}    & \textbf{30.57}   & -2.82             & \textbf{12.14}   & \cellcolor{lg}43.20 & \cellcolor{lg}61.94 & \cellcolor{lg}41.67 & \cellcolor{lg}16.96 &  23.03 & 15.59 & 39.09 & 64.02 & 73.54 \\
\midrule
Shared Bottom                  & Equal                                     & 5.27             & 24.44            & -9.96             & 6.58             & \cellcolor{lg}40.33 & \cellcolor{lg}60.05 & \cellcolor{lg}45.93 & \cellcolor{lg}18.70 &  24.48 & 17.51 & 35.12 & 61.29 & 71.32 \\
\bottomrule

\end{tabular}
\end{table}

\begin{table}
\caption[Performance comparison of SAAL and baseline methods for the Cityscapes dataset.]{Performance comparison of SAAL and baseline methods for the Cityscapes dataset. The MTL method with the best relative task improvement is bolded. Individual task metrics that outperform the STL baseline are highlighted in grey.}\label{table cityscapes}
\centering
\fontsize{6pt}{7pt} \selectfont
\begin{tabular}{ll|rrrr|rrrrr}
\toprule
\multicolumn{1}{l}{}                   &                           & \multicolumn{4}{c}{Relative Improvement $\uparrow$} & \multicolumn{2}{c}{\textbf{S}emantic Seg. $\uparrow$} & \multicolumn{2}{c}{\textbf{P}art Seg. $\uparrow$} & \multicolumn{1}{c}{\textbf{D}isp. $\downarrow$} \\
\cmidrule(l){3-6}
\cmidrule(l){7-8}
\cmidrule(l){9-10}
\cmidrule(l){11-11}
\multicolumn{1}{l}{Architecture}       & Method        & $\Delta_{\text{S}}$ & $\Delta_{\text{P}}$  & $\Delta_{\text{D}}$             & $\Delta_{\text{MTL}}$ & mIoU       & Pix acc                     & mIoU     & Pix acc     & Rel err                        \\
\midrule
STL                            & -                                         & 0.00          & 0.00          & 0.00          & 0.00         & 53.95              & 80.35              & 52.99              & 97.59              & 85.36 \\
\midrule
\multirow{10}{*}{Half-shared}  & Equal                                     & -6.48         & -3.25         & 16.13         & 2.13         & 47.64              & 79.34              & 49.62              & 97.46              & \cellcolor{lg}71.59 \\
                               & Uncertainty~\cite{MTL_weight_uncertainty} & -3.21         & \textbf{-0.10}& 17.73         & 4.81         & 50.90              & 79.74              & 52.88              & 97.59              & \cellcolor{lg}70.22 \\
                               & DWA~\cite{MTL_attention}                  & -5.24         & -3.92         & 14.48         & 1.77         & 48.98              & 79.32              & 48.93              & 97.42              & \cellcolor{lg}73.00 \\
                               & Auto-$\lambda$~\cite{MTL_auto_lambda}     & -4.06         & -2.01         & 16.60         & 3.51         & 50.12              & 79.53              & 50.90              & 97.52              & \cellcolor{lg}71.19 \\
                               & PCGrad~\cite{MTL_PCGrad}                  & -6.35         & -3.05         & 16.41         & 2.34         & 47.79              & 79.32              & 49.83              & 97.45              & \cellcolor{lg}71.36 \\
                               & CAGrad~\cite{MTL_PCGrad}                  & -4.61         & -1.52         & 16.65         & 3.51         & 49.50              & 79.56              & 51.40              & 97.55              & \cellcolor{lg}71.15 \\
                               & GradDrop~\cite{MTL_conflict_averse}       & -7.11         & -3.91         & 14.02         & 1.00         & 47.12              & 79.09              & 48.95              & 97.39              & \cellcolor{lg}73.39 \\
                               & SAAL$_\text{e}$                           & -2.12         & -0.90         & 17.89         & \textbf{4.96}& 52.01              & 79.84              & 52.04              & 97.58              & \cellcolor{lg}70.09 \\
                               & SAAL$_\text{w}$                           & -3.54         & -2.55         & \textbf{18.39}& 4.11         & 50.54              & 79.75              & 50.34              & 97.52              & \cellcolor{lg}69.67 \\
                               & SAAL$_\text{ew}$                          & -3.04         & -0.56         & 18.13         & 4.85         & 51.00              & 79.87              & 52.39              & \cellcolor{lg}97.61& \cellcolor{lg}69.88 \\
\midrule
Shared Bottom                  & Equal                                     & \textbf{-1.69}& -2.11         & 5.82          & 0.67         & 52.35              & 80.02              & 50.82              & 97.47              & \cellcolor{lg}80.39 \\
\bottomrule          
\end{tabular}
\end{table}

\begin{table}
\caption[Performance comparison of SAAL and baseline methods for the CelebA dataset.]{Performance comparison of SAAL and baseline methods for the CelebA dataset. The MTL method with the best relative task improvement is bolded. Individual task metrics that outperform the STL baseline are highlighted in grey.}\label{table celeba}
\centering
\fontsize{6pt}{7pt} \selectfont
\begin{tabular}{ll|r|rrrrrrrrr}
\toprule
                                         &                           & \multicolumn{1}{c}{Rel. Imp.} & \multicolumn{1}{c}{A1} & \multicolumn{1}{c}{A2} & \multicolumn{1}{c}{A3} & \multicolumn{1}{c}{A4} & \multicolumn{1}{c}{A5} & \multicolumn{1}{c}{A6} & \multicolumn{1}{c}{A7} & \multicolumn{1}{c}{A8} & \multicolumn{1}{c}{A9} \\
Architecture                             & Method        & $\Delta_{\text{MTL}} \uparrow$ & acc                       & acc                       & acc                       & acc                       & acc                       & acc                       & acc                       & acc                       & acc                                           \\
\midrule
STL                            & -                                         & 0.00          & 91.70              & 79.17              &               76.55& 82.42              & 98.25              & 94.38              & 68.03              & 80.95              & 85.58 \\
\midrule
\multirow{10}{*}{Half-shared}  & Equal                                     & -0.32         & \cellcolor{lg}92.09& 78.32              & \cellcolor{lg}77.47& 81.57              & \cellcolor{lg}98.27& 94.05              & \cellcolor{lg}68.39& 78.71              & \cellcolor{lg}85.89 \\
                               & Uncertainty~\cite{MTL_weight_uncertainty} & -0.30         & \cellcolor{lg}91.82& 77.85              & \cellcolor{lg}77.68& 81.60              & \cellcolor{lg}98.31& 93.18              & \cellcolor{lg}68.08& 80.30              & \cellcolor{lg}86.04 \\
                               & DWA~\cite{MTL_attention}                  & 0.17          & \cellcolor{lg}92.00& 78.58              & \cellcolor{lg}77.85& 81.65              & \cellcolor{lg}98.31& \cellcolor{lg}94.46& \cellcolor{lg}68.52& 80.29              & \cellcolor{lg}86.79 \\
                               & Auto-$\lambda$~\cite{MTL_auto_lambda}     & -0.70         & 90.48              & 76.53              & \cellcolor{lg}77.44& 81.75              & 97.90              & 92.52              & \cellcolor{lg}68.11& 80.90              & \cellcolor{lg}86.03 \\
                               & PCGrad~\cite{MTL_PCGrad}                  & 0.33          & \cellcolor{lg}92.05& 78.99              & \cellcolor{lg}77.92& 81.89              & \cellcolor{lg}98.34& \cellcolor{lg}94.72& \cellcolor{lg}68.44& 80.78              & \cellcolor{lg}86.50 \\
                               & CAGrad~\cite{MTL_PCGrad}                  & -0.30         & \cellcolor{lg}91.97& 78.46              & \cellcolor{lg}77.06& 80.88              & 98.23              & 94.37              & 68.00              & 79.69              & \cellcolor{lg}86.37 \\
                               & GradDrop~\cite{MTL_conflict_averse}       & 0.06          & \cellcolor{lg}91.93& 78.33              & \cellcolor{lg}77.75& 81.50              & \cellcolor{lg}98.30& 94.32              & \cellcolor{lg}68.40& 80.74              & \cellcolor{lg}86.29 \\
                               & SAAL$_\text{e}$                           & 0.24          & 91.68              & 78.51              & \cellcolor{lg}77.95& 82.13              & \cellcolor{lg}98.26& 94.31              & \cellcolor{lg}68.85& 80.30              & \cellcolor{lg}86.56 \\
                               & SAAL$_\text{w}$                           & \textbf{0.76} & 91.14              & \cellcolor{lg}79.29& \cellcolor{lg}78.91& 81.72              & 98.20              & 94.00              & \cellcolor{lg}69.28& \cellcolor{lg}81.39& \cellcolor{lg}86.38 \\
                               & SAAL$_\text{ew}$                          & 0.63          & 91.44              & 78.91              & \cellcolor{lg}78.05& 81.91              & 98.25              & 94.15              & \cellcolor{lg}68.39& 80.46              & \cellcolor{lg}86.53 \\
\midrule
Shared Bottom                  & Equal                                     & -0.06         & 91.58              & 79.17              & \cellcolor{lg}78.16& 81.61              & 98.18              & 94.04              & \cellcolor{lg}69.00& 80.47              & 84.08 \\
\bottomrule
\end{tabular}
\end{table}

\subsection{Runtime Comparison} 
Table~\ref{table asym time} presents a comparative analysis of the average runtime for each optimisation method. Among the baseline methods, Auto-$\lambda$ had the slowest runtime. The other loss weighting strategies, Uncertainty and DWA, had minimal increases in computation time over Equal, while the gradient manipulation strategies had moderate increases in computation time. Overall, SAAL$_\text{w}$ was the slowest method, and SAAL$_\text{ew}$ was the second slowest due to the exclusion of self-auxiliaries with negative transfer. The extended runtime of these methods, along with Auto-$\lambda$, can be attributed to their task coefficient learning strategy, which necessitates multiple forward and backward passes per batch. SAAL$_\text{e}$ additionally requires pre-computing of actual task relationships. All SAAL strategies were slower on datasets with more tasks, due to the quadratic scaling of candidate self-auxiliaries. 

Developing techniques that exploit asymmetric task relationships in MTL models without significant computational overhead may improve the effectiveness of these systems in diverse applications. An approach may be to rely on transferring learning signals from primary tasks, removing the need to train self-auxiliary tasks and avoiding redundancy in learning signals in network components trained by multiple cloned tasks. Improved strategies for learning task relationships, particularly asymmetric relationships, may also reduce computational costs.

\begin{table}[t]
\caption[Runtime comparison of multi-task optimisation methods.]{Runtime comparison of multi-task optimisation methods. We display the average runtime for a training batch for each method, relative to the Equal weighting method.}\label{table asym time}
\centering
\fontsize{6pt}{7pt} \selectfont
\begin{tabular}{l|rrr}
\toprule
            & NYUv2 & Cityscapes & CelebA \\
\midrule
Equal                              & \textit{1.00} & \textit{1.00} & \textit{1.00}  \\
Uncertainty~\cite{MTL_weight_uncertainty} & 1.01          & 1.01          & 1.04  \\
DWA~\cite{MTL_attention}                  & 1.00          & 1.00          & 1.00   \\
Auto-$\lambda$~\cite{MTL_auto_lambda}     & 4.24          & 4.03          & 3.82  \\
PCGrad~\cite{MTL_PCGrad}                  & 1.31          & 1.26          & 1.46  \\
CAGrad~\cite{MTL_PCGrad}                  & 1.15          & 1.13          & 1.50  \\
GradDrop~\cite{MTL_conflict_averse}       & 1.33          & 1.10          & 1.46  \\
SAAL$_\text{e}$                           & 1.41          & 1.64          & 1.54  \\
SAAL$_\text{w}$                           & 6.84          & 6.55          & 8.30  \\
SAAL$_\text{ew}$                          & 4.87          & 5.27          & 5.86  \\
\bottomrule
\end{tabular}
\end{table}

\subsection{Task Relationships}
\begin{figure}[t]
    \centering
    \begin{subfigure}[t]{0.27\textwidth}
        \vspace{0pt}
        \includegraphics[width=\textwidth]{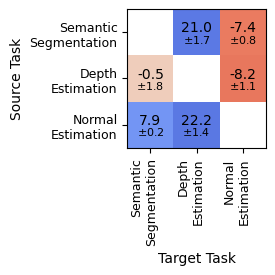}
        \caption{NYUv2}
        \vskip 0.05in
        \includegraphics[width=\textwidth]{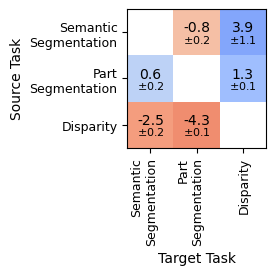}
        \caption{Cityscapes}
    \end{subfigure}
    \hspace{0.006\textwidth}
    \begin{subfigure}[t]{0.7\textwidth}
        \vspace{0pt}
        \includegraphics[width=\textwidth]{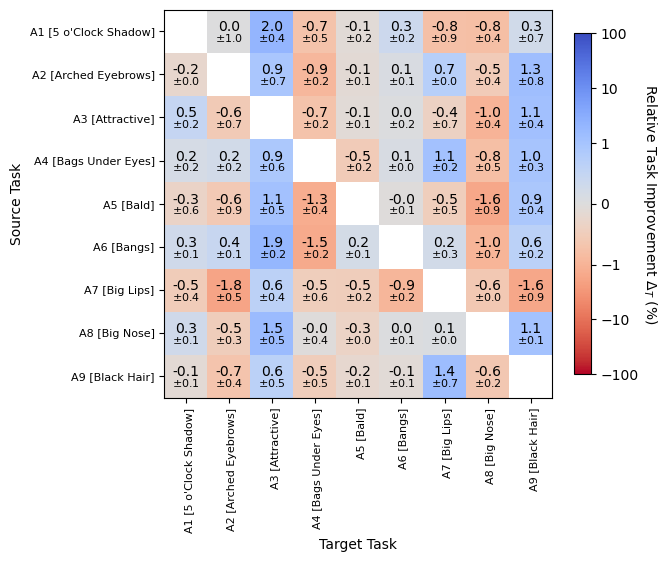}
        %\vskip 0.32in
        \caption{CelebA}
    \end{subfigure}
    \caption{Visualisation of task relationships for computer vision datasets.}
    \label{fig task relos computer vision}
\end{figure}

We illustrate actual task relationships of these datasets in Figure~\ref{fig task relos computer vision}. Task relationships are determined by computing the average relative task improvement for a Shared Bottom model over an STL baseline for each combination of tasks, identically to our task enumeration strategy.
%, where all task relationships with positive asymmetric transfer being included as a self-auxiliary in the training of SAAL$_\text{e}$ and SAAL$_\text{ew}$.

Consistent with previous results, the NYUv2 and Cityscape datasets contain asymmetric task relationships with our evaluation setup, with NYUv2 exhibiting more important relationships. This may explain the greater gain in performance from our method when applied to NYUv2 compared to Cityscapes. We note that task relationships share some consistencies across these two datasets, with both datasets exhibiting positive knowledge transfer from instance segmentation tasks to depth or disparity (inverse depth) but either no or negative knowledge transfer in the other direction, consistent with prior works~\cite{MTL_task_groupings,MTL_taskonomy}. 

The CelebA dataset also contains asymmetric task relationships. Many tasks exhibit positive knowledge transfer towards predicting A3 (whether a person is attractive), while many tasks exhibit negative knowledge transfer towards predicting A8 (whether a person has a big nose). We speculate that this may be because predicting attractiveness is an abstract concept contingent on diverse facial attributes while predicting nose size is a specific concept that may not share visual features in common with other facial attributes. The relationship between these tasks is asymmetric as A8 helps A3 learn but not vice versa. Other asymmetric relationships without intuitive explanations exist, such as between A7 (whether a person has big lips) and A9 (whether a person has black hair).

Overall, task relationships appear to be more influenced by the target task than the source task, with all datasets producing more consistent task relationships with varying source tasks than with varying target tasks. This characteristic may explain the prevalence of asymmetric task relationships in these datasets. Previous empirical studies on task relationships measure the average knowledge transfer in both directions, obscuring this result~\cite{MTL_task_groupings,MTL_task_relationship_RSA,MTL_task_relationship_DDS}.

% TASK RELATIONSHIPS
\subsection{Task Relationship Learning} 
We investigate the effectiveness of our asymmetric task relationship learning strategy used in SAAL$_\text{w}$, comparing it to some common task relationship learning approaches discussed in Section~\ref{asym related works}. We evaluate the effectiveness of task relationship learning strategies against the actual task relationships displayed in Figures~\ref{fig task relos computer vision}, then compute the average Spearman rank correlation with task relationship learning methods for each target task~\cite{MTL_task_relationship_DDS,MTL_task_relationship_RSA}. We include the following approaches:

\paragraph{Look-ahead loss.} During the training of a Shared Bottom MTL model, a virtual training step is applied using the loss term from each task, and the change in training and validation loss of all other tasks is computed. We include the median lookahead loss between tasks over a training run.    

\paragraph{Gradient angle.} We compute the median angle between the gradients of different tasks' training signals over a training run.

\paragraph{Feature similarity.} To measure feature similarity between models trained on different tasks, we evaluate the transferability of feature representations between tasks. An STL model is trained on a source task, frozen and fine-tuned on a target task, allowing us to compute the predictive power of the latent representation learnt by each task applied to other tasks. We use a transfer function between the encoder and decoder to increase flexibility in the feature transfer, using two convolutional layers following existing methodology~\cite{MTL_taskonomy}. Unlike Standley et al.~\cite{MTL_task_groupings}, we investigate asymmetric task relationships rather than averaging both transfer directions.
%This allows us to avoid relying on feature similarity metrics that approximate task relationships~\cite{MTL_task_relationship_DDS,MTL_task_relationship_RSA,MTL_task_relationship_attribution} 
    
\paragraph{SAAL$_\text{w}$ coefficients.} The SAAL$_\text{w}$ coefficients $\omega$ at the end of training, which aim to estimate the importance of knowledge transfer between each task pair. \\

\begin{table}[t]
\caption[Spearman rank correlation coefficients of actual task relationships vs. task relationship learning approaches.]{Spearman rank correlation coefficients of actual task relationships vs. task relationship learning approaches. The approach with the strongest correlation is
bolded.}\label{table task relationships}
\centering
\fontsize{6pt}{7pt} \selectfont
\begin{tabular}{l|rrrrr}
\toprule
                            & NYUv2          & Cityscapes     & CelebA    \\
\midrule
Look-ahead$_{\text{train}}$ & -0.33          & -0.33          & -0.06     \\
Look-ahead$_{\text{val}}$   & -0.33          & -0.33          & -0.06     \\
Gradient angle              & -0.33          & -0.33          & -0.05     \\
Feature Similarity          & \textbf{1.00}  & \textbf{1.00}  & -0.07     \\
SAAL$_\text{w}$ coefficients & \textbf{1.00}  & \textbf{1.00}  & \textbf{0.26} \\
\bottomrule
\end{tabular}
\end{table}

We observe that look-ahead loss and gradient angle correlate poorly with actual task relationships for all datasets. SAAL$_\text{w}$ coefficients and feature similarity were perfectly correlated with the actual task relationships of the NYUv2 and Cityscapes datasets, but correlated more poorly on CelebA. These results may be influenced by NYUv2 and Cityscapes having fewer tasks, simplifying the task relationship learning process. These results indicate that the loss weighting strategy employed by SAAL$_\text{w}$ can learn task coefficients that correlate to actual task relationships.

\section{Conclusion}\label{section asym conclusion}
%We investigated the understudied of asymmetric knowledge transfer in multi-task learning and 
%to avoid 
We proposed Self-Auxiliary Asymmetric Learning, the first multi-task optimisation strategy that can induce asymmetric knowledge transfer during training. This allows multi-task learning models to exploit asymmetric task relationships and mitigate directed negative transfer. Our method induces transfer by training auxiliary task clones using model parameters specific to other tasks, using several strategies to determine which task clones to include in training. We showed that many MTL datasets have asymmetric task relationships and demonstrated that our method outperforms many existing MTL optimisation methods on common MTL computer vision datasets.

\bibliography{iclr2021_conference}
\bibliographystyle{unsrtnat}
%\bibliographystyle{splncs04}
%\bibliographystyle{iclr2021_conference}

%\appendix
%\section{Appendix}
%You may include other additional sections here.

\end{document}